\documentclass[pdflatex,sn-vancouver-num]{sn-jnl}

\usepackage{graphicx}%
\usepackage{multirow}%
\usepackage{amsmath,amssymb,amsfonts}%
\usepackage{amsthm}%
\usepackage{mathrsfs}%
\usepackage[title]{appendix}%
\usepackage{xcolor}%
\usepackage{textcomp}%
\usepackage{manyfoot}%
\usepackage{booktabs}%
\usepackage{algorithm}%
\usepackage{algorithmicx}%
\usepackage{algpseudocode}%
\usepackage{listings}%

\theoremstyle{thmstyleone}%
\theoremstyle{thmstyletwo}%

\theoremstyle{thmstylethree}%

\begin{document}

\title[Reinforcement Learning and Machine Ethics: A Survey]{Reinforcement Learning and Machine Ethics: A Survey}


\author*[1]{\fnm{Ajay} \sur{Vishwanath}}\email{ajay.vishwanath@uia.no}

\author[2]{\fnm{Louise} \sur{Dennis}}\email{louise.dennis@manchester.ac.uk}

\author[3]{\fnm{Marija} \sur{Slavkovik}}\email{marija.slavkovik@uib.no}

\affil*[1]{\orgname{University of Agder}, \country{Norway}}

\affil[2]{\orgname{University of Manchester}, \country{United Kingdom}}

\affil[3]{ \orgname{University of Bergen}, \country{Norway}}


\abstract{Machine ethics is the field that studies how ethical behaviour can be accomplished by autonomous systems. While there exist some systematic reviews aiming to consolidate the state of the art in machine ethics prior to 2020, these tend to not include work that uses reinforcement learning agents as entities whose ethical behaviour is to be achieved. The reason for this is that only in the last years have we witnessed an increase in machine ethics studies within reinforcement learning. We present here a systematic review of reinforcement learning for machine ethics and machine ethics within reinforcement learning. Additionally, we highlight trends in terms of ethics specifications, components and frameworks of reinforcement learning, and environments used to result in ethical behaviour. Our systematic review aims to consolidate the work in machine ethics and reinforcement learning, thus filling the gap in the state of the art machine ethics landscape.}




\maketitle

\section{Introduction} \label{intro}

Machine ethics ``is concerned with the behaviour of machines towards human users and other machines'' \cite{Anderson2007}.  This includes studying how to implement machine reasoning in a way that includes the consideration of ethical factors \cite{Anderson11,Wallach08}. Since the 2006 AAAI Workshop on Machine Ethics~\cite{Anderson2007}, there has been a rapid expansion in proposed systems and formalisms for performing machine ethics (e.g, ~\cite{Vanderelst18,Dennis2016,Bringsjord2006,Bench-Capon2017,LiaoPST23,GrandiLPA22})  as evident from surveys of the field~\cite{nallur20,tolmeijer20,Yu18}.  

Reviews of machine ethics, at present, cover the period up to 2020. Currently, we observe a notable trend in reinforcement learning, specifically focusing on the ethical behaviour of reinforcement learning agents.

In reinforcement learning (RL) \cite{Sutton2018} an artificial agent makes decisions in an environment and receives rewards and punishments as outcomes of those decisions. The goal, typically of the learning process, is to find a policy on how to make decisions so that an optimal reward is attained. RL, and specifically deep RL that combines neural networks and RL,  has been of particular recent  interest  since its successful application in training artificial agents to play human-level computer games \cite{mnih_human-level_2015}. Since RL is about training an artificial agent in how to make decisions; it is very intuitive as to why ethical decision-making would receive attention in the RL community.

Early papers on the RL and machine ethics topic, such as \cite{armstrong2015motivated}, gained recognition, along with subsequent work by researchers ~\cite{ijcai2018p843,Hendrycks21,nahian21,DBLP:conf/ijcai/Rodriguez-SotoL21,neufeld_enforcing_2022,tennant2023modeling,DBLP:journals/aiethics/VishwanathBGMO23}.  There is a gap in the understanding of the state of the art in machine ethics if that understanding does not include the most recent work on RL. Our goal is to bridge this gap by conducting a literature review on RL in the context of machine ethics.  While our motivation derives from the observation that much work in this sub-field has taken place since the survey articles of 2020, we included earlier works in our survey in order to present a complete view of the application of RL to machine ethics, including important early results.

We used a systematic review approach similar to that outlined in~\cite{systematic_review}.  We defined a review protocol that specifies a research question and the methods used to perform the review; we defined and documented our search strategy; we defined the inclusion and exclusion criteria; and we specified the information to be obtained from each paper identified by the process.



Following a keywords search in the usual repositories of artificial intelligence research articles, we found 808 potential papers which we have narrowed down to 71 papers that were on topic and which were carefully analysed. We analysed the papers based on the RL framework they deploy, how they modify which RL ``component" to achieve ethical behaviour, and which ethical theory they directly or indirectly implement, as well as what kind of implementation examples the papers use. Beyond this, we observe and discuss the trends in this body of work and draw recommendations for future research in this field. 

This article contributes to both the disciplines of RL and machine ethics by establishing a clear state of the art of the intersection of these two fields and giving future work a good anchor to position and grow from existing experience. Furthermore, we highlight the challenges and possible pitfalls of RL and machine ethics, thus ensuring a responsible and sustainable development of what at present appears to be a trendy and growing research field.

\section{Background and related work}
In this section, we discuss RL and the various algorithms employed in the papers we review. This is followed by a discussion on previous work in machine ethics and relevant gaps.

\subsection{Reinforcement Learning} \label{RL_basics}
Reinforcement learning (RL) is a machine learning framework in which an agent navigates an environment with states $s \in S$, actions $a \in A$, next states $s' \in S'$ and rewards $r \in R$. Such an agent learns through trial and error an optimum behaviour which maximises a numerical reward function \cite{Sutton2018}. This optimum behaviour is denoted by $\pi^*$, or the optimum policy. Markov decision processes (MDP) are a mathematically idealised form of the RL problem. In an MDP, the probability of each possible value of the next state $s'$ and reward $r$ depends on the immediately preceding state and action, $s$ and $a$, completely characterise the environment and that these states, actions, and rewards are finite. An important conundrum in RL is the exploration-exploitation trade-off, where if the agent performs over-exploration, the agent may not converge to an optimum, while if the agent over-exploits, the agent could potentially waste time on poor actions and fail to consider better alternatives \cite{Sutton2018}.

A method to solve MDPs is using value function approximation, where a numerical value is calculated for being in a state $V(s)$ and committing an action in a given state $Q(s, a)$. As the agent visits different states in the environment, a value table is updated based on the reward obtained. This method is often referred to as Q-learning, where a Q table is updated and based on these Q-values an optimal policy is determined. However, an issue with such an approach is that in an environment with a very high state space, such as the game of chess, it can be very expensive to compute a Q-table. 

In high state-space scenarios, a neural network is often used to encode the state space and output the Q-value for being in that state. This framework of using a neural network in the Q-learning algorithm is referred to as Deep Q-learning. There are, however, limitations to using value function approximation, since there is an exploration-exploitation conundrum. These networks are said to be more exploratory in nature and often fail to converge to an optimum \cite{Sutton2018}. Such behaviour can be exacerbated in environments with high stochasticity or vast state spaces, where the balance between continual exploration and fine-tuning of previously learned policies becomes increasingly challenging. To overcome this issue, actor-critic methods are used where two neural networks, an actor and a critic, are used to predict action probabilities and Q-values, respectively, instead of only calculating Q-values. Methods such as Proximal Policy Optimization (PPO), Deep Deterministic Policy Gradient (DDPG), etc. are popular actor-critic methods.

There are other types of RL, such as inverse RL, multi-objective RL and co-operative RL. In inverse RL (IRL) \cite{AbeelNg}, the agent learns the reward functions in an environment, given the policy. This is useful in applications such as robotics, where the robot can "watch" a human demonstrate a task (optimal policy) and then the robot can imitate the same, while learning a reward function in its environment. Next, multi-objective RL (MORL), an agent is trained to optimise conflicting objective functions, demonstrated in works by Rodriguez et al. \cite{rodriguez2022instilling} and several others discussed in this review. Finally, co-operative inverse RL is a variant of IRL in a multi-agent scenario. However, unlike classical IRL, where the human is the basis for learning, the other agents in the system often perform active teaching, learning and communicative actions \cite{hadfield-menell_cooperative_2016}. Other RL frameworks such as safe/constrained RL are often employed to prevent hazardous failures in RL algorithms during training and deployment \cite{Gu2023}. While both terminologies are sometimes used interchangeably, constrained RL is a subset of safe RL that formally defines constraints and maximises rewards while meeting constraints.

\subsection{Moral philosophy}
\textit{Ethics} being the crucial piece of the machine ethics puzzle, is succinctly summarised here. The three major ethical theories in moral philosophy are: deontology, consequentialist and virtue ethics. A consequentialist \cite{tolmeijer20} centers their morality on the outcome of an action, rather than the action itself. Utilitarianism is a type of consequentialist theory, where an action that maximises utility, which is a metric that quantifies the outcome, is deemed morally praiseworthy. There are two types of utilitarians, act- and rule-utilitarians. While act-utilitarians scrutinise every act to ensure they maximise the overall well-being based on the utility principle, rule-utilitarians prioritise the rules which strictly maximise utility, irrespective of whether it is an \textit{optimal} action. There are other variants of utilitarianism and consequentialist theories. However, for the purposes of this paper, which focuses on RL-based implementations, it would suffice to keep in mind that consequentialist ethics places a majority of the onus on the consequences.

Deontological ethics, as opposed to consequentialist theories, scrutinises an act based on whether it fulfilled a set of rules or duties. Deontology derives from Kantian ethics, who said, ``So act that you use humanity, whether in your own person or in the person of any other, always at the same time as an end, never merely as a means.'' \cite{kant_groundwork_2012}. There are several variants of deontology as well, and in \ref{ethical_theory} where we discuss our results, we delineate some of them based on the RL implementations. The third major theory is virtue ethics, which is a normative theory that centers on excellence of moral character. For example, \textit{temperance} is regarded as a virtue where we seek a balance between overindulgence in pleasure and avoidance of pleasures. This balance is often referred to as the \textit{golden mean}. Then there is \textit{prima-facie} reasoning where duties such as fidelity, gratitude, etc., are central and decisions are made based on the context. This is different from deontology in the sense that duties in \textit{prima-facie} reasoning are derived from human relations and experiences, rather than rational principles and categorical imperatives, as is the case in deontology. With this basic knowledge\footnote{In the context of RL used in machine ethics, the moral theories are predominantly based on Western moral philosophy. However, we acknowledge that several there exist moral theories derived from Chinese, Indian, and African schools of thought, and it is plausible to implement them in reinforcement learning algorithms.}, we discuss next some of the systematic reviews in machine ethics.

\subsection{Systematic reviews in machine ethics}
We briefly summarise the findings of the three reviews in machine ethics, their findings and methodology.

Tolmeijer et al. \cite{tolmeijer20} focused on the implementations of machine ethics specifically, and found 49 papers  from Web of Science, Scopus, ACM Digital Library, Wiley Online Library, ScienceDirect, AAAI Publications, Springer Link, and IEEE Xplore. These papers were then analysed based on the types of ethical theory they implemented, the nontechnical aspects when implementing those theories, and technological details. 

Nallur \cite{nallur20} considers the landscape of machine ethics implementations, focusing on the different problems, techniques, methodologies and evaluation methodologies that are used. He did not do a systematic review, but rather a more qualitative analysis of selected prominent works. 

Yu et al. \cite{Yu18} survey advances in techniques for incorporating ethics into AI. They consider publications in   ``leading AI research conferences including AAAI, AAMAS, ECAI and IJCAI, as well as articles from well-known journals''. They organise the papers into four categories:  Exploring Ethical Dilemmas, Individual Ethical Decision Frameworks, Collective Ethical Decision Frameworks; and 4. Ethics in Human-AI Interactions. There is, of course, a significant overlap among the papers included in all three surveys.

Cervantes et al. \cite{cervantes_artificial_2020} performed a survey by first categorising artificial moral agents based on Moor's definition \cite{Moor2006}: implicit ethical agents, explicit ethical agents, and full ethical agents. Implicit agents are those whose actions are restricted to avoid unethical consequences, while explicit ethical agents follow ethical rules crafted explicitly for a purpose. Full ethical agents are those that are capable of reasoning and making ethical decisions like humans. Cervantes et al. analysed individual contributions in depth and highlighted strengths and weaknesses. Moreover, they categorised agents based on their approach to learning ethical principles: top-down (ethical principles pre-specified), bottom-up (infer ethical rules from their environment), and hybrid approaches (a combination of top-down and bottom-up). While not being a systematic literature review, Cervantes et al. highlighted features from examples based on a concise taxonomy. 

We here follow to a large degree the approach of Tolmeijer et al. \cite{tolmeijer20} in terms of where we search for papers. This is because Tolmeijer et al. \cite{tolmeijer20} is the only systematic review in machine ethics. 
Understanding the state of the art in machine ethics is not without challenges. The papers published in this area are sent to a wide spectrum of venues. Machine ethics is also not a very large area of research, and it is not very precisely positioned in the context of the AI alignment and AI Ethics initiatives. Therefore, it has to be accepted that any systematic review of the field is not likely to be exhaustive, even if we limit ourselves to a specific domain or methodology such as RL.

\section{Survey methodology}
In this section, we describe the research questions, the strategy used to find papers related to machine ethics and RL, following which we discuss the process used to review the found articles.

\subsection{Research questions}
Using the following formulated research questions (RQ), we plan the review procedure.
\begin{description}
    \item[RQ1:] The first research question pertains to the RL algorithm and its corresponding components, such as the type of RL framework (MORL, DQN, etc.) adopted, the components of RL modified (state space, action space, policy function, value function, and reward function), and the RL environment (examples). Capturing these distinctions enables us to uncover trends in the field that researchers are inclined towards, for the development of artificial moral agents.
    \item[RQ2:] This question studies the type of ethical theories (deontology, consequentialism, etc.) realised by the RL agents and who decides the ethics. A strength of RL algorithms is to optimise an agent to achieve a certain goal. This research question offers insights into which ethical theories can be programmed and why researchers gravitate towards a certain theory. Moreover, understanding who decides what is ethical in a given context enables us to identify gaps in the field.
    \item[RQ3:] Which research contributions are prevalent in literature? Depending on how far a field has matured, it will be interesting to find out the proportion of purely theoretical, empirical and mathematical contributions.
\end{description}

We address RQ1, a portion of RQ2, and RQ3 in Section \ref{analysis}, and the remainder of RQ2 in \ref{discussion}. 

\subsection{Search strategy}

Our review on implementations in machine ethics focuses on RL, hence we mandated this terminology while querying the databases. Since RL is to be used towards machine ethics implementations, we used alternative search terms based on seminal contributions in machine ethics, such as ``artificial morality'' \cite{Allen05,cervantes_artificial_2020}, ``machine morality'' and ``robot ethics'' \cite{Malle2016}, ``computational ethics'' \cite{anderson2006approach}, ``roboethics'' \cite{coeckelbergh2009personal}, ``artificial moral agents'' \cite{cervantes_artificial_2020}, ``moral agent'' \cite{tennant2023modeling}, ``value alignment'' \cite{rodriguez2022instilling},  ``normative reasoning'' \cite{kasenberg_norms_2018}, and ``ethical dilemma'' \cite{Abel2016ReinforcementLA}. These terms have been used interchangeably to communicate the embedding of ethical theories into machines/computers/artificial intelligence agents. We show the query used below:

\begin{center}
``reinforcement learning'' AND ( ``machine ethics'' OR ``artificial morality'' OR ``moral agent'' OR ``machine morality'' OR ``computational ethics'' OR ``roboethics'' OR ``robot ethics'' OR ``artificial moral agents'' OR ``value alignment'' OR ``normative reasoning'' OR ``ethical dilemma'')
\end{center}

The query was used to search the following databases (Table \ref{tab:databases}). In the right-hand side column we give  the number of `hits' from the query. Further filtering was performed on these `hits'.  
\begin{table}[ht]
    \centering
    \begin{tabular}{|l|c|}
        \hline
        Database & Number of results \\
        \hline
        ACM Digital Library & 174 \\
        IEEE Xplore & 54 \\
        ScienceDirect & 78 \\
        Scopus & 161 \\
        SpringerLink & 251 \\
        Web of Science & 28 \\
        Wiley Online Library & 175 \\
        \hline
        \textbf{Total} & \textbf{921}\\
        \hline
    \end{tabular}
    \caption{Databases and respective number of results for the search query used.}
    \label{tab:databases}
\end{table}

\subsection{Selection process}

\begin{figure}[ht]
    \centering
    \includegraphics[width=12cm]{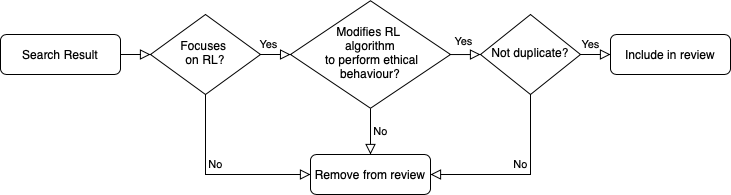}
    \caption{The selection process used to eliminate irrelevant articles but keep the relevant ones. First is to check whether the article focuses on RL, followed by checking if the RL algorithm is used for moral decision-making, and finally to eliminate duplicates that appeared across multiple databases.}
    \label{fig:selection-process}
\end{figure}

A total of 921 papers were further filtered based on relevance to RL and machine ethics. We included only journal articles, conference papers, and book chapters which were published before 4th January 2024. The diagram in Figure~\ref{fig:selection-process} shows the elimination process we used to arrive at the final list of manuscripts.

After applying these selection criteria, we ended up with 87 papers for review. Subsequently, we conducted a more focused review by filtering the papers based on their relevance to both machine ethics and RL, resulting in a total of 59 articles. Some examples of papers excluded from the 87 papers are \cite{butlin_ai_2021,hadfield-menell_cooperative_2016,serramia18}. For example, Hadfield-Menell et al. \cite{hadfield-menell_cooperative_2016} specifically examined the issue of value alignment in conjunction with reinforcement learning methodologies, approaching the problem from a robotics-centric paradigm. The term ``values'', as articulated by these researchers, pertains to the imperative for robotic systems to exhibit behaviours that are congruent with established human decision-making protocols, as opposed to adhering to a specific ethical framework or moral philosophy. On the other hand, Serramia et al. \cite{serramia18}, while making ``reinforcement learning'' references, do not explicitly modify an existing RL algorithm to encode ethics. In the next section, we delve deeper into the taxonomy creation and review process.

\section{The analysis} \label{analysis}
In this section, we discuss the taxonomy and categorisations used to organise our survey based on research questions, RQ1-RQ3 (except for a portion of RQ2 which resides in Discussion (Section \ref{discussion}). 

\subsection{Component of RL Modified}
There are several aspects that affect RL algorithms and their performance on tasks. For instance, the challenge of designing a good reward function to correctly incentivise agents can make or break the agent's performance. Other aspects such as policy functions, value functions, etc., also play a part. In terms of developing RL algorithms which make more ethical choices, several researchers modify either one or more of these components to achieve the desired result. We categorised the research articles based on which component of RL was modified, and we present our findings in Table \ref{tab:component}. 

\begin{table}[ht]
    \centering  \begin{tabular}{|c|l|}
    \hline
        RL component & \\
        \hline
        Reward function & \cite{alizadeh_alamdari_be_2022, Alcaraz2023, peschl_training_2021, sun_designing_2018, Wu_Lin_2018, alkoby_teaching_2019, vamplew_human-aligned_2018, Rodriguez-Soto2020, sanneman_validating_2023, byrd_learning_2022, chaput_learning_2023, Noothigattu2019TeachingAA, tassella_artificial_2023, rosello-marin_ethical_2022, gu_approach_2022, rodriguez2022instilling, haas_moral_2020, chaput_multi-agent_2021, butlin_ai_2021, stenseke_artificial_2022, rolf_social_2018, DBLP:conf/ijcai/Rodriguez-SotoL21,tennant2023modeling,riedl_using_2016,Abel2016ReinforcementLA,herlau_moral_2022,gao_human_2023,ajmeri_fleur_2022} \\
        Policy function & \cite{robinson_action_2023, slater_golem_2021, Noothigattu2019TeachingAA, ijcai2018p843, rosello-marin_ethical_2022, kaas_raising_2021, DBLP:conf/ijcai/Rodriguez-SotoL21, amato_pragmatic-pedagogic_2020}\\
        Value function & \cite{alizadeh_alamdari_be_2022, pan_rewards_2023, shea-blymyer_generating_2022, balkenius_outline_2016} \\
        Others & \cite{balakrishnan_incorporating_2019, malvone_admissible_2023, plebe_neurocomputational_2015,ammanabrolu-etal-2022-aligning,kasenberg_norm_2018,fagundes_design_2016,ajmeri_centralized_2022,platzer_normative_2021,aydogan_normative_2023,DBLP:journals/aiethics/VishwanathBGMO23, Kasenberg2018} \\
    \hline
    \end{tabular}
    \caption{Component of RL modified.}
    \label{tab:component}
\end{table}

We can observe from these groupings that the majority of the articles modify either the reward function or the policy function. Reward functions act as a means to instill ethical behaviour, since the agent can obtain a higher reward or incur a lower penalty if it chooses the ethical option. On the other hand, inverse RL algorithms are typically used to modify the policy function through demonstrations of an expert policy, based on which the agent learns the optimal reward function. Also, a few researchers demonstrated ethical behaviour preference in RL agents by modifying either the value function or the neural network's objective function (see Table \ref{tab:component}). Other components include the action space \cite{ammanabrolu-etal-2022-aligning}, neural network \cite{plebe_neurocomputational_2015} in the case of deep RL algorithms, path planning \cite{malvone_admissible_2023} and regret \cite{balakrishnan_incorporating_2019}.

\subsection{RL framework}
Similar to observing the component of the RL algorithm modified, it was also interesting to categorise and analyse the RL \textit{framework} adopted to bring about ethical behaviour. By \textit{framework}, we refer to the type of RL used based on which the relevant component is modified. We briefly introduced some of the RL frameworks in Section \ref{RL_basics}, such as inverse RL, multi-objective RL, constrained RL, etc. Each of these frameworks has variants and sub-variants, but these variations within frameworks largely function in a similar fashion. In Table \ref{tab:framework}, we specify the frameworks and the corresponding research works.
\begin{table}[ht]
    \centering \begin{tabular}{|c|l|}
    \hline
    RL framework &  \\
    \hline
        Regular RL & \cite{stenseke_artificial_2022, Alcaraz2023, guarda_machine_2024, pinka_synthetic_2021, crook_anatomy_2021, arnold_value_2017, neufeld_enforcing_2022, Chen2019, krening_q-learning_2023, Wu_Lin_2018, alizadeh_alamdari_be_2022, Kasenberg2018, shea-blymyer_generating_2022, balakrishnan_incorporating_2019, tennant2023modeling, riedl_using_2016, balkenius_outline_2016,Stenseke2021,ajmeri_fleur_2022,platzer_normative_2021,ajmeri_centralized_2022,fagundes_design_2016,herlau_moral_2022,kasenberg_norm_2018} \\
       Multi-objective RL  & \cite{Rodriguez-Soto2020, rosello-marin_ethical_2022, tassella_artificial_2023, RodriguezSoto2022, rodriguez2021guaranteeing, vamplew_human-aligned_2018, rodriguez2022instilling, chaput_learning_2023, haas_moral_2020, Peschl2022, DBLP:conf/ijcai/Rodriguez-SotoL21, Noothigattu2019TeachingAA, alkoby_teaching_2019} \\
        Inverse RL and variants & \cite{robinson_action_2023, Kasenberg2018, amato_pragmatic-pedagogic_2020, kaas_raising_2021, alkoby_teaching_2019, slater_golem_2021, peschl_training_2021}\\
        Constrained/safe RL & \cite{byrd_learning_2022, ijcai2018p843, hoshi_what_2018,aydogan_normative_2023}\\
        Other RL algorithms & \cite{rolf_social_2018, plebe_neurocomputational_2015, DBLP:journals/aiethics/VishwanathBGMO23, sun_designing_2018,chaput_multi-agent_2021,ammanabrolu-etal-2022-aligning,Abel2016ReinforcementLA,gao_human_2023} \\
         \hline
    \end{tabular}
    \caption{RL frameworks used to result in ethical behaviour.}
    \label{tab:framework}
\end{table}

Most of the researchers adopted multi-objective RL (MORL) and inverse RL (IRL). The advantage of using MORL is that ethics specifications can be codified as part of the reward function, in addition to the main objective. An MORL agent navigates its environment, maximising its reward, thereby fulfilling its main tasks along with ethical sub-tasks, or in many cases, completes its main task in an ethical manner. While in IRL, the agent learns an optimal reward function based on an ethical policy demonstrated by an expert. Variations of IRL include co-operative IRL, and adversarial IRL, which use human experts as co-operative and adversarial means, respectively, rather than an expert teacher, to embed ethical behaviour in an artificial agent.

Other techniques, such as constrained RL and safe RL, teach an agent to ethically navigate an environment by avoiding unethical states or policies. Apart from these RL frameworks, several researchers have also adopted \textit{regular} RL, which refers to RL algorithms such as multi-arm bandits, deep Q-learning, actor-critic models, etc., which most applications of RL use. Finally, other RL algorithms include dynamic self-organising maps \cite{chaput_multi-agent_2021}, POMDP \cite{Abel2016ReinforcementLA}, multi-agent RL \cite{sun_designing_2018}, social RL \cite{rolf_social_2018}, implicit RL \cite{plebe_neurocomputational_2015} and affinity-based RL \cite{DBLP:journals/aiethics/VishwanathBGMO23}. We also observed several uses of multi-agent RL, across these frameworks; however, for simplicity's sake, we did not single out this categorisation.

\subsection{Implementation Example}

A wide variety of settings were used for experimental evaluation of systems. Very few of these settings were shared across teams of researchers, revealing a lack of comparative benchmarking sets.  As a result, ethical frameworks were generally compared against baselines in which there was no attempt to represent ethics, rather than alternative frameworks or theories.  A broad categorisation of these examples can be seen in Table~\ref{tab:example}.

The only family of examples that appeared in various forms across multiple studies and teams of researchers involved the equitable sharing of resources in a multi-agent setting -- in which the metrics used were the ability of the community to support all its members.  These scenarios ranged from abstract Grid Worlds representing foraging games with various opportunities to share and/or steal resources from other agents, to more realistic settings based on, for instance, water distribution among several communities in~\citet{malvone_admissible_2023}.

Another common family of examples was abstract representations of forbidden behaviour--often as a Grid World with forbidden areas. \citet{alkoby_teaching_2019} has a variant on this that involved the agent learning to play Tetris taking into account human preferences about colour placement.  In many cases, it was difficult to see these examples as specifically about ethical behaviour, as opposed to more general examples around \textit{safe} RL, learning preferences, or constrained learning.

Beyond these two distinct families of examples, there was a varied range of examples which involved motion planning in an environment that contained conflicting values or values conflicting with goals.  For instance, two papers, ~\citet{neufeld_enforcing_2022} and \citet{Noothigattu2019TeachingAA}, required an agent to learn to play vegan pac-man where it was deemed unethical to eat the ghosts.  Another set of papers, all by the same team~\citet{Rodriguez-Soto2020,RodriguezSoto2022,rodriguez2021guaranteeing,rodriguez2022instilling} involved a ``public civility game'' in which agents need to reach a goal which is obstructed by rubbish, and they must learn to divert to place the rubbish in a bin, rather than throwing it at the other agents involved.

Two examples involved moderately sophisticated simulations of smart grids (\citet{chaput_multi-agent_2021}) and financial markets (\citet{byrd_learning_2022}).

The remaining cases were ad-hoc examples, some with clear ethical relevance -- e.g., around patient consent to treatment (\citet{kasenberg_norms_2018}).  However, in some cases, particularly in systems where the focus was primarily on learning human norms, rather than a more specialised focus on values, we saw examples that seemed more about learning preferences than anything with particular ethical force -- for instance \citet{amato_pragmatic-pedagogic_2020} focuses on robot chef learning to cook meals that its owner will like.

\begin{table}[htbp]
    \centering
    \begin{tabular}{|p{.475\linewidth}|p{.375\linewidth}|}
    \hline
    Implementation example & \\
    \hline
      Equity Example & \cite{stenseke_artificial_2022,Stenseke2021,sun_designing_2018,haas_moral_2020,DBLP:conf/ijcai/Rodriguez-SotoL21,malvone_admissible_2023,crook_anatomy_2021,ajmeri_centralized_2022}\\
       Abstract Example with Forbidden/Preferred States & \cite{Kasenberg2018,haas_moral_2020,plebe_neurocomputational_2015,sanneman_validating_2023,arnold_value_2017,alkoby_teaching_2019,kasenberg_norm_2018,aydogan_normative_2023} \\
       Motion Planning with conflicting goals and/or values  &  \cite{Wu_Lin_2018,robinson_action_2023,neufeld_enforcing_2022,Noothigattu2019TeachingAA,Rodriguez-Soto2020,RodriguezSoto2022,rodriguez2021guaranteeing,rodriguez2022instilling,Peschl2022,peschl_training_2021,platzer_normative_2021,fagundes_design_2016,herlau_moral_2022} \\
       Sophisticated Simulations & \cite{chaput_multi-agent_2021,byrd_learning_2022} \\
         Other & \cite{gu_approach_2022,rosello-marin_ethical_2022,alizadeh_alamdari_be_2022,pan_rewards_2023,balakrishnan_incorporating_2019,balakrishnan2019,kasenberg_norms_2018,amato_pragmatic-pedagogic_2020,Chen2019,krening_q-learning_2023,Abel2016ReinforcementLA,ajmeri_fleur_2022,gao_human_2023} \\
       \hline
    \end{tabular}
    \caption{Categorisation of examples.}
    \label{tab:example}
\end{table}

\subsection{Ethical Theory} \label{ethical_theory}
A crucial aspect to explore in this systematic literature review is the ethical theories used by the researchers in their respective works. We reviewed each article to try to understand its underlying ethical proclivities. Based on this examination, we found that the majority do not explicitly state their ethics specifications and, due to this, it was challenging to distinguish these manuscripts based on moral philosophy. However, to help researchers better understand the ethical underpinnings behind these machine ethics papers, we focus on the ethical theory that influences the RL algorithm the most, rather than all the theories at play. For example, if researchers codified social norms and then optimised a utility function as is done in RL, then we consider such implementations as deontological. On the other hand, if a paper does not explicitly mention deontological codification based on rules or norms, then it is considered consequentialist (which also includes variants of utilitarianism). We also found several other researchers using human values, where they optimised metrics based on human evaluation. In Table \ref{tab:ethics}, we summarise the ethics distinctions.

\begin{table}[ht]
    \centering
    \begin{tabular}{|c|l|}
    \hline
    Ethical theory & \\
    \hline
       Consequentialist  &  \cite{Rodriguez-Soto2020, rosello-marin_ethical_2022, sun_designing_2018, pan_rewards_2023, rodriguez2021guaranteeing, haas_moral_2020, shea-blymyer_generating_2022, vamplew_human-aligned_2018, haas_moral_2020, Peschl2022, Chen2019, krening_q-learning_2023,tennant2023modeling,Abel2016ReinforcementLA,herlau_moral_2022,ajmeri_centralized_2022}\\
       Deontology & \cite{chaput_multi-agent_2021, gu_approach_2022, RodriguezSoto2022, neufeld_enforcing_2022, rodriguez2022instilling, Kasenberg2018, byrd_learning_2022,ammanabrolu-etal-2022-aligning, DBLP:conf/ijcai/Rodriguez-SotoL21, kasenberg_norms_2018, arnold_value_2017, hoshi_what_2018,tennant2023modeling,aydogan_normative_2023,platzer_normative_2021,fagundes_design_2016,kasenberg_norm_2018} \\
       Virtue ethics & \cite{Stenseke2021, DBLP:journals/aiethics/VishwanathBGMO23, crook_anatomy_2021, Ribeiro2016,tennant2023modeling}\\
       Human expert values & \cite{Wu_Lin_2018, chaput_multi-agent_2021, robinson_action_2023, tassella_artificial_2023, Noothigattu2019TeachingAA, chaput_learning_2023, amato_pragmatic-pedagogic_2020, kaas_raising_2021, alkoby_teaching_2019, slater_golem_2021, Peschl2022, ijcai2018p843, sanneman_validating_2023,riedl_using_2016} \\
       Others (unspecified) & \cite{alizadeh_alamdari_be_2022, Alcaraz2023, rolf_social_2018, plebe_neurocomputational_2015, pinka_synthetic_2021, balkenius_outline_2016,gao_human_2023}\\
       \hline
    \end{tabular}
    \caption{Ethics specification used by researchers.}
    \label{tab:ethics}
\end{table}

From this it can be observed that a majority adopted a consequentialist, deontological and human expert approach. Deontological and consequentialist ethics-based approaches being dominant is not surprising \cite{Stenseke2021, tolmeijer20}, since they are arguably easier to codify and implement (compared to, for example, virtue ethics). We also see that there are several works under "human expert values", and that these "values" have been codified to help RL agents make ethical decisions in their environment. In an upcoming section, we deep-dive into who decides these values. Next, virtue ethics has fewer implementations, since the codification is challenging. However, it has been argued as a promising direction to pursue \cite{DBLP:journals/aiethics/VishwanathBGMO23, stenseke_artificial_2022}. Finally, the remaining manuscripts did not explicitly specify and instead had other goals such as social compliance \cite{rolf_social_2018}, emulated imagination \cite{pinka_synthetic_2021}, etc.

\subsection{Contribution type}
We found several differences in claimed types of contributions in our literature review. Some were position papers, which were argumentative in nature, theorizing on how RL could be used in machine ethics research. Others were purely empirical contributions, either by referring to a previously proposed theory on a new problem, or improving upon existing results. The remaining ones were a combination of argumentative, empirical and mathematical proofs. It was interesting to explore these contribution types mainly to understand the maturity of the field: RL in machine ethics. The contributions are summarised in Table \ref{tab:contribution_type}.
\begin{table}[ht]
    \centering
     
    \begin{tabular}{|p{.325\linewidth}|p{.425\linewidth}|}
    \hline
    Contribution type & \\
    \hline
       Argumentative/Theoretical  & \cite{robinson_action_2023, ammanabrolu-etal-2022-aligning, tassella_artificial_2023, guarda_machine_2024, haas_moral_2020, kaas_raising_2021, pinka_synthetic_2021, crook_anatomy_2021, slater_golem_2021, DBLP:journals/aiethics/VishwanathBGMO23, sanneman_validating_2023, hoshi_what_2018, chaput_multi-agent_2021, stenseke_artificial_2022, Stenseke2021, sun_designing_2018, balakrishnan_incorporating_2019, chaput_learning_2023, DBLP:conf/ijcai/Rodriguez-SotoL21, plebe_neurocomputational_2015, amato_pragmatic-pedagogic_2020, peschl_training_2021, rolf_social_2018, vamplew_human-aligned_2018, rodriguez2022instilling, balkenius_outline_2016,aydogan_normative_2023,gao_human_2023,platzer_normative_2021,ajmeri_centralized_2022,kasenberg_norm_2018,Abel2016ReinforcementLA} \\
       Empirical & \cite{Wu_Lin_2018, rosello-marin_ethical_2022, alizadeh_alamdari_be_2022, pan_rewards_2023, rodriguez2021guaranteeing, Kasenberg2018, byrd_learning_2022, Peschl2022, kasenberg_norms_2018, malvone_admissible_2023, Chen2019, krening_q-learning_2023, Noothigattu2019TeachingAA, alkoby_teaching_2019, chaput_multi-agent_2021, stenseke_artificial_2022, Stenseke2021, sun_designing_2018, balakrishnan_incorporating_2019, chaput_learning_2023, DBLP:conf/ijcai/Rodriguez-SotoL21, plebe_neurocomputational_2015, amato_pragmatic-pedagogic_2020, peschl_training_2021, neufeld_enforcing_2022, shea-blymyer_generating_2022, ijcai2018p843, rodriguez2022instilling,tennant2023modeling,riedl_using_2016,ajmeri_fleur_2022,gao_human_2023,platzer_normative_2021,ajmeri_centralized_2022,fagundes_design_2016,herlau_moral_2022,Abel2016ReinforcementLA} \\
       Math. Proof/Formalisation & \cite{rolf_social_2018, vamplew_human-aligned_2018, neufeld_enforcing_2022, shea-blymyer_generating_2022, ijcai2018p843, rodriguez2022instilling,aydogan_normative_2023,kasenberg_norm_2018,Abel2016ReinforcementLA} \\
       \hline
    \end{tabular}
    \caption{Type of research.}
    \label{tab:contribution_type}
\end{table}

Usually, new fields begin with purely argumentative and position papers, discussing ways in which the research could move forward. Based on this, new mathematical and empirical evaluations are performed, and newer positions are formed. We can observe from the Table \ref{tab:contribution_type} that empirical evaluations form a large part of the research while still having a sizeable number of theoretical contributions. In a later section, we discuss the research field as a whole and the associated trends based on contribution type.

\section{Discussion} \label{discussion}
We here discuss the trends we observed in the surveyed papers and draw out some recommendations for future research based on our observations. We conclude this section by analysing research question RQ2 related to the human-factors responsible for deciding the ethics.

\subsection{Recent increase in} 
\begin{figure}[ht]
    \centering
    \includegraphics[width=0.45\textwidth]{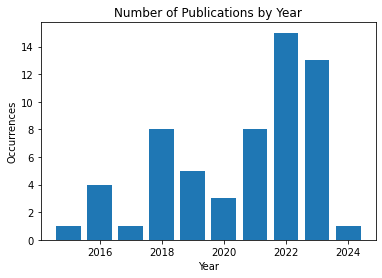}
    \caption{Recent increase in contributions in RL based on the papers chosen for review.}
    \label{fig:RL-trend}
    \hspace{-0.5cm}
\end{figure}

contributions in RL
Based on our survey, we found a steady increase in the use of RL in machine ethics. Figure \ref{fig:RL-trend} illustrates the trend, where we plot the number of contributions per year. It is clear from this bar chart that, since 2020, the attention towards ME and RL has seen a consistent rise, with a peak in 2022.

\subsection{No moral theories}

Wallach \cite{Wallach08} categorises approaches to the problem of machine ethics as either ``top-down'' or ``bottom up''.  Broadly, top-down approaches seek to operationalise some ethical theory from philosophy and apply this to a decision faced by the machine.  Bottom-up approaches seek to learn ethical behaviour from data.  

Much of the work covered in the survey papers~\cite{nallur20,tolmeijer20,Yu18} and performed in the early years of the field has taken a top-down approach with a few systems taking a hybrid line -- for instance, the GenEth system~\cite{AndersonA2018} in which inputs from ethics experts were used to enable a system to learn explicit rules for differentiating between choices in a medical ethics situation. In RL, the trend is not to follow a specific moral theory that will inform the agent's policy. As we reflected in the previous section, there is no explicit intention to have RL agents trained into following a specific moral theory.

 \subsection{Paradigm trends}
 A significant number of approaches we encountered have encoded the ethics in some fashion within the reward or objective function. Within these we observed three trends:
\begin{description}
    \item[Ethical RL is Multi-Objective RL] In many cases, the ethics was implemented as a reward signal in addition to other (non-ethical) rewards, such as maximising profit or efficiency.  This frames ethical RL as a form of multi-objective RL.
    \item[Ethics as Constraints] In other cases, the ethics were represented as constraints.  Ethical behaviour does not confer any reward upon the agent, although in some cases unethical behaviour incurred a penalty.  This frames ethical RL as a form of constrained RL.  In some cases, there is an additional requirement that behaviour remain ethical during training, in which case ethical RL becomes a form of Safe RL.
    \item[Multi-agent RL] Another significant theme, highlighted by the number of implementation examples based around equitable distribution of resources, is framing ethical behaviour in terms of maximising group benefit in a multi-agent context.
\end{description}
We would argue that it is a valuable contribution to the field to point out that much by way of ethical behaviour can be achieved by appropriate application of techniques from Constrained Multi-Objective RL and Safe RL, but that going forward researchers should take this point as established and, if working in these areas, focus their attention on how ethics can be represented in terms of these multiple objectives and constraints.

\subsection{Human factors: whose ethics?}


Based on our analysis of ethics specifications and moral philosophy, we found that several researchers have included a human factor to implement ethical behaviour in RL agents. It is either the developer, the user, an expert, or an adversary (Table \ref{tab:human}), whose ethics an RL agent must try and emulate. Clearly, there is no real agreement on which ethical theory we, as humanity, fall back on. We use  some version of utilitarianism, social contracts, virtue ethics, deontology, and other ethical theories interchangeably.

\begin{table}[ht]
    
    \begin{tabular}{|p{.4\linewidth}|p{.425\linewidth}|}
    \hline
    Whose ethics? & \\
    \hline
       Designer/developer/researcher &  \cite{balakrishnan_incorporating_2019,balakrishnan2019,rosello-marin_ethical_2022,Kasenberg2018,plebe_neurocomputational_2015,arnold_value_2017,DBLP:conf/ijcai/Rodriguez-SotoL21,sun_designing_2018,stenseke_artificial_2022,crook_anatomy_2021,malvone_admissible_2023,alizadeh_alamdari_be_2022,haas_moral_2020,Chen2019,Wu_Lin_2018,Stenseke2021,Rodriguez-Soto2020,RodriguezSoto2022,rodriguez2021guaranteeing,rodriguez2022instilling,Peschl2022,kasenberg_norms_2018,byrd_learning_2022,chaput_multi-agent_2021,pan_rewards_2023,robinson_action_2023,neufeld_enforcing_2022,Alcaraz2023,vamplew_human-aligned_2018,pinka_synthetic_2021,DBLP:journals/aiethics/VishwanathBGMO23,hoshi_what_2018,tennant2023modeling,ajmeri_fleur_2022,aydogan_normative_2023,platzer_normative_2021,ajmeri_centralized_2022,fagundes_design_2016,herlau_moral_2022,kasenberg_norm_2018,Abel2016ReinforcementLA}\\
       Algorithm user & \cite{tassella_artificial_2023, chaput_learning_2023} \\
       Co-operative human & \cite{amato_pragmatic-pedagogic_2020, slater_golem_2021}\\
       Human expert/demonstrator & \cite{kaas_raising_2021, Noothigattu2019TeachingAA, peschl_training_2021, riedl_using_2016} \\
       Human-machine team & \cite{krening_q-learning_2023, sanneman_validating_2023, alkoby_teaching_2019} \\
       Others (unspecified) & \cite{gu_approach_2022, rolf_social_2018, Ribeiro2016, balkenius_outline_2016,gao_human_2023}\\
       \hline
    \end{tabular}
    \caption{Who decides the ethics?}
    \label{tab:human}
\end{table}
The problem of not clearly subscribing to a well-understood moral theory is the human affinity for bias. 
Human bias is a hazard for designing applications that affect thousands of people who have their own plethora of biases that may be counter to the algorithm developer's biases. Hence, it is crucial that algorithms consider a variety of viewpoints and evolve with changing times. While it is unrealistic to expect developers/product managers/researchers to know and understand so many viewpoints, it is still important that the makers strive towards at least an abstract understanding while developing ethical RL algorithms. Most of the algorithms in our survey have not explicitly specified who decides the ethics or right from wrong. However, we urge the research community to explicitly identify and report their ethical biases/assumptions made in their implementations, since future researchers who extend their work might be vulnerable to propagating the same \cite{bai2022traininghelpfulharmlessassistant}.

Some works (Table \ref{tab:human}), where the human designer/developer/researcher decides ethical behaviour, do indeed explicitly mention the human factor, where the RL algorithms are trained based on whether certain results are allowed or not. For instance, some of this is demonstrated using ethical movie recommendations, where certain movies are excluded from the results for underage users \cite{balakrishnan2019}. Others \cite{chaput_multi-agent_2021} use a \textit{judging} RL agent trained with ethical values to train other RL agents in the environment. An important challenge for such works is to codify human values and translate these into the algorithm. In cases where an expert human demonstrator is involved \cite{kaas_raising_2021} in training the algorithm, it's important that they are aware of competing ethical perspectives rather than training the model on one specific moral behaviour. In the case of using co-operative humans and a human teammate, these are useful ways to train AI, but it's also vital to infer what exactly is learnt by the model in the form of explanations and interpretations. Lastly, an interesting approach is to let the algorithm \textit{user} specify the ethics \cite{tassella_artificial_2023}. This shifts the responsibility from the developer to the user, who can then customise the algorithm to their ethical needs. However, this approach could also affect the usability of the algorithm because, with so much customisation, the interaction experience might become troublesome for the user. Hence, striking the balance of usability and customisation becomes important.

\section{Summary} \label{summary}
We presented and analysed the results of a systematic review of articles that use RL for machine ethics or consider the problem of machine ethics in RL.  The increased volume of papers in machine ethics RL in recent years points to a promising area of research.
The main challenge for the RL in machine ethics community going forward is to organise itself around a taxonomy that will help authors position their efforts against existing work. 

The secondary challenge is to avoid ``anecdotal ethics''. RL in machine ethics most of the time uses bottom-up and sometimes hybrid approaches (in the sense of \cite{Wallach08}) to achieve artificial moral agency, to avoid developing methods around the moral sensitivities and judgments of its developers. The secondary challenge is thus to establish a firmer connection with the fields of moral philosophy and moral psychology \cite{Bello2023}. The systematic evaluation and benchmarking of the abilities of reinforcement learning ethical agents remains an open problem. 

\backmatter

\section*{Declarations}
The authors declare no conflict of interest.


\bibliography{sn-bibliography}

\end{document}